\documentclass[paper]{ieice}
\usepackage[pdftex]{graphicx,xcolor}
\usepackage{cite}
\usepackage{algorithmic}
\usepackage{textcomp}
\usepackage{colortbl}
\usepackage[fleqn]{amsmath}
\usepackage{amssymb,amsfonts}
\usepackage[varg]{newtxmath}
\newcommand{\bi}[1]{\ensuremath{\boldsymbol{#1}}}
\newcommand{\T}{\ensuremath{^{\text{T}}}}

\usepackage{tabularx,booktabs}
\newcolumntype{C}{>{\centering\arraybackslash}X} 
\newcolumntype{R}{>{\raggedright\arraybackslash}X}
\newcommand{\multicell}[2]{\setlength{\tabcolsep}{-0.5mm} \begin{tabular}{c}
                                                              #1\\#2
\end{tabular}}

\field{D}
\vol{106}
\no{4}
\jalcdoi{DOI: 10.1587/transinf.2022EDP7200}
\title{CAMRI Loss: Improving the Recall of a Specific Class without Sacrificing Accuracy}
\titlenote{This paper is an extension of the authors' previous work presented at 2022 International Joint Conference on Neural Networks (IJCNN 2022).}
\authorlist{%
    \authorentry[nishiyama@mdl.cs.tsukuba.ac.jp]{Daiki Nishiyama}{n}{1,3}\MembershipNumber{}
    \authorentry{Kazuto Fukuchi}{n}{2,3}\MembershipNumber{}
    \authorentry{Youhei Akimoto}{n}{2,3}\MembershipNumber{}
    \authorentry{Jun Sakuma}{m}{2,3}\MembershipNumber{1310828}
}
\affiliate[1]{Master's Program in Computer Science, Degree Programs in Systems and Information Engineering, Graduate School of Science and Technology, University of Tsukuba}
\affiliate[2]{Department of Computer Science, Faculty of Engineering, Information and Systems, University of Tsukuba}
\affiliate[3]{RIKEN Center for Advanced Intelligence Project}

\received{2022}{11}{15}
\onlinefirst{2023}{1}{23}

\begin{document}
    \maketitle
    \begin{summary}
        In real world applications of multiclass classification models, misclassification in an important class (e.g., stop sign) can be significantly more harmful than in other classes (e.g., no parking).
        Thus, it is crucial to improve the recall of an important class while maintaining overall accuracy.
        For this problem, we found that improving the separation of important classes relative to other classes in the feature space is effective.
        Existing methods that give a class-sensitive penalty for cross-entropy loss do not improve the separation.
        Moreover, the methods designed to improve separations between all classes are unsuitable for our purpose because they do not consider the important classes.
        To achieve the separation, we propose a loss function that explicitly gives loss for the feature space, called class-sensitive additive angular margin (CAMRI) loss.
        CAMRI loss is expected to reduce the variance of an important class due to the addition of a penalty to the angle between the important class features and the corresponding weight vectors in the feature space.
        In addition, concentrating the penalty on only the important class hardly sacrifices separating the other classes.
        Experiments on CIFAR-10, GTSRB, and AwA2 showed that CAMRI loss could improve the recall of a specific class without sacrificing accuracy.
        In particular, compared with GTSRB's second-worst class recall when trained with cross-entropy loss, CAMRI loss improved recall by 9\%.
    \end{summary}

    \section{Introduction}
    \label{sec:introduction}
    Recently, the generalization accuracy of classification prediction using deep convolutional neural networks (DCNNs) has significantly improved\cite{lecun1998gradient, krizhevsky2012imagenet, lecun2015deep},
    and DCNNs are expected to be applied to various fields such as automated driving\cite{yurtsever2020survey} and medical diagnosis support\cite{bakator2018deep, saranya2019survey}.
    However, if the trained classification model misclassifies in the real world, it may cause serious accidents.

    We consider the multiclass classification problem of identifying road signs in automatic driving.
    If the model of an automatic vehicle misclassifies a "Stop" sign as some other sign, the vehicle will enter the intersection without pausing, likely causing a traffic accident.
    Meanwhile, misclassifying a "No Parking" sign as a "No Parking Daytime" sign is relatively unlikely to cause accidents or other harm.
    Therefore, the low recall of classes such as "Stop" and "No Entry," which cause actual harm due to misclassification, is a particularly important problem.

    Thus, to use a DCNN multiclass classification model in the real world, there are cases where we want to increase the recall of a particularly important class.
    Nevertheless, the accuracy should not deteriorate at the cost of improving the recall of a particular class.
    Therefore, it is necessary to find a technique to improve the recall of an important class without sacrificing the accuracy of other classes.

    \subsection{Related Work}
    \label{sec:related}
    Threshold tuning\cite{margineantu2000bootstrap}
    is considered a naive method for the considered problem.
    However, this method improves the recall of a particular class but sacrifices that of other classes, deteriorating the accuracy.
    In other words, because it only adjusts the tradeoff ratio between the accuracy and recall of a particular class, it is impossible to both improve recall and maintain accuracy.

    Several methods employ cost-sensitive learning\cite{Elkan01thefoundations,kukar1998cost}.
    Cost-sensitive learning is a method for predicting classification with a loss function that imposes a relatively large cost (i.e., misclassification cost) on unacceptable errors.
    Several methods\cite{panchapagesan2016multi,aurelio2019learning,ho2019real,frogner2015learning}
    consider the importance of classes that need to improve recall as a penalty and use this penalty as a loss function that weights cross-entropy (CE) loss.
    However, as shown in Fig.~\ref{tab:1},
    these methods do not always sufficiently improve the recall of an important class.
    Even if the recall can be improved,
    the accuracy may be impaired (Fig.~\ref{fig:5}).
    This phenomenon is because these methods cannot improve the separation of an important class in the feature space.
    The details are discussed in Section~\ref{sec:analysis}.

    For our goal, we need to increase the separation between an important class and other classes while maintaining the separation between unimportant classes.
    To achieve such separation, the neural network should be trained to generate features that (1) the feature vectors of the important class are well-isolated and localized from the other classes and (2) the feature vectors of each class are well-separated from each other.
    Recently, various loss functions\cite{liu2016large,liu2017sphereface,liang2017soft,wang2018additive,wang2018cosface,deng2019arcface} have been developed to achieve this.
    These methods consider classification by cosine similarity between the representative vector of each class and the feature vector.
    The vectors are obtained by projecting them onto the unit hypersphere\cite{zhai2018classification,ranjan2017l2}.
    With this setup, the margin between classes in the feature space is expected to increase.
    Then, the cosine term is manipulated with some constant factor so that the separation between classes is improved, which is called the \textit{margin}.
    Liu \textit{et al.}\cite{liu2016large,liu2017sphereface} proposed a method to introduce an angular margin loss (SphereFace) between the class corresponding to the feature vector and other classes to promote the expansion of interclass variance.
    Liang \textit{et al.}\cite{liang2017soft} and Wang \textit{et al.}\cite{wang2018additive,wang2018cosface} proposed an additive margin loss (CosFace) to stabilize the optimization.
    Deng \textit{et al.}\cite{deng2019arcface} proposed an additive angular margin loss (ArcFace) that can be interpreted geometrically, and ArcFace has reported good performance in face recognition tasks.

    Although these angular-based loss functions can increase the separability of all classes, they do not concentrate on improving the separability of a specific class.
    In other words, these methods are unsuitable for our purposes because they cannot concentrate on improving the recall of an important class.

    \subsection{Our Contribution}
    The intuitive approach to improve the recall of an important class is to optimize the CE loss by assigning a larger weight to the loss for the important class.
    However, as shown in Section~\ref{sec:experiment}, this approach does not attain our purpose even with intensive weight parameter tuning because of the sacrifice of accuracy.
    In addition, existing approaches to increase the separability of all classes are unsuitable because they cannot concentrate on improving the recall of an important class.
    In our research, we found that improving the separation of important classes relative to other classes in the feature space is necessary for our purpose.
    Our major contributions are as follows.

    \begin{itemize}
        \item We propose a loss function explicitly giving loss for the feature space to improve recall of an important class without sacrificing accuracy, called class-sensitive additive angular margin (CAMRI) loss.
        CAMRI loss adds the margin, called \textit{class-sensitive angular margin}, to the angle between the feature vector and the weight vector corresponding to the important class of the last fully connected (FC) layer only when the feature is labeled with the important class.
        CAMRI loss makes the feature distribution of the important class more compact than that of other classes and improves the separability of the features labeled with the important class from other classes in the feature space.
        \item We empirically showed that CAMRI loss improves the recall of the important class without sacrificing accuracy.
        We performed experiments using three datasets: Cooperative Institute For Alaska Research 10 (CIFAR-10)\cite{torralba200880}, German Traffic Sign Recognition Benchmark (GTSRB)\cite{Stallkamp-IJCNN-2011}, and Animals with Attributes 2 (AwA2)\cite{8413121}.
        In eight of nine cases, CAMRI loss achieved a higher value of recall improvement than the existing method while maintaining accuracy.
        \item We experimentally showed two advantages of CAMRI loss.
        First, CAMRI loss can improve the recall of any class; the degree of recall improvement depends on the difficulty of learning class features and baseline recall obtained using CE loss.
        Second, CAMRI loss can improve each recall of multiple important classes without sacrificing accuracy compared with CE loss.
    \end{itemize}

    The remainder of this article is organized as follows.
    In Section~\ref{sec:pre}, we formulate the multiclass classification problem and related loss functions.
    In Section~\ref{sec:analysis}, we analyze the characteristics of related loss functions based on their contours in the feature space. Then, we propose a method to improve the separation of features labeled with an important class by adding the class-sensitive angular margin for the important class.
    In Section~\ref{sec:experiment}, we experimentally evaluate the improvement in recalls of our method and related methods using multiple datasets and discuss the characteristics of our method through additional experiments.
    Finally, we conclude this study in Section~\ref{sec:conclusion}.

    \section{Preliminaries}
    \label{sec:pre}
    We represent the $i$th element of vector \bi{a} as $a_i$, the $(i,j)$th element of matrix \bi{A} as $A_{i,j}$,
    the $i$th row of matrix \bi{A} as $\bi{A}_{i}$, and the $j$th column of matrix \bi{A} as $\bi{A}_{:,j}$.

    \subsection{Multiclass Classification}
    Let $K$ be the number of classes, $\mathcal{X}$ be the input space, $\mathcal{Y}=\left\{ 1,\cdots,K \right\}$ be the output space,
    and $f\colon\mathcal{X}\rightarrow\mathcal{Y}$ be the multiclass classification model.
    We use a convolutional neural network (CNN) for $f$.
    We train the CNN to represent the feature vector $\bi{z}_n\in \mathbb{R}^D$ with $D$ dimensions,
    using a pair of input images and teacher labels $\left\{ \left( \bi{x}_n,t_n \right)\right\}_{n=1}^N$ for training.
    The one hot representation of $t_n$ is given by $\bi{y}_n$,
    $\bi{z}_n$ is given to the FC layer with weights $\bi{W}\in\mathbb{R}^{D\times K}$ and bias $\bi{b}\in\mathbb{R}^K$,
    and the FC layer outputs $\bi{o}_n=\bi{W}\T \bi{z}_n + \bi{b}$.
    The softmax function transforms $\bi{o}_n$ into a probability vector $\bi{h}_n$,
    whose elements are interpreted as the predicted probabilities corresponding to each class.
    The CE loss is often used for classifier training, which takes $\bi{h}_n$ and $\bi{y}_n$ as inputs.
    The inference of the model $f$ is given by $\text{argmax}_{k \in \mathcal{Y}} h_{n,k}$.

    In this study, our purpose is to improve the recall of a specific class without sacrificing accuracy in a multiclass classification model compared with CE loss.

    \subsection{Related Loss Functions}
    In this section, we introduce several types of loss functions related to our proposal.
    Now, let class $\kappa$ be an important class.

    \subsubsection{CE Loss}
    First, we formulate the commonly used CE loss.
    When the number of samples is $N$ and the number of classes is $K$, the CE loss in multiclass classification is given by
    \begin{equation}
        \mathcal{L}_{\mathrm{ce}}=-\frac{1}{N} \sum_{n=1}^{N} \sum_{k=1}^{K}\left(y_{n, k} \log \left(h_{n, k}\right)\right).
        \label{eq:1}
    \end{equation}

    \subsubsection{Weighted CE (WCE) Loss}
    Second, we introduce WCE loss\cite{panchapagesan2016multi,aurelio2019learning,ho2019real} as the method that imposes class-specific penalties on CE loss.
    Let $w_k$ be the weight parameter of class $k$.
    By penalizing the loss with $w_k$ for each class, WCE loss is given by
    \begin{equation}
        \mathcal{L}_{\text{wc}}=
        -\frac{1}{N} \sum_{n=1}^{N} \sum_{k=1}^K
        \left( w_{k} y_{n,k} \log \left(h_{n,k}\right)  \right).
        \label{eq:2}
    \end{equation}
    For our goal, we should set the $\kappa$th of \bi{w} to a value greater than $1$ and set the other elements to $1$.

    \subsubsection{Categorical real world WCE (CRWWCE) loss}
    Ho \textit{et al.}\cite{ho2019real} introduced real world WCE (RWWCE) loss, which penalizes each class's false negatives and false positives independently with different costs.
    The multiclass classification extension of RWWCE loss, termed CRWWCE loss, is defined by
    \begin{eqnarray}
        \mathcal{L}_{\text{rw}}&=&
        -\frac{1}{N} \sum_{n=1}^{N}\sum_{k=1}^{K}
        \Bigl( c^{\text{fn}}_{k} y_{n,k} \log \left(h_{n,k}\right)  \nonumber\\
        && \mbox{\ \ \ } +\sum_{k^{\prime} \neq k} C^{\text{fp}}_{k^{\prime},k}  y_{n,k} \log \left(1-h_{n,k^{\prime}}\right) \Bigr).
        \label{eq:3}
    \end{eqnarray}
    $\bi{c}^{\text{fn}}\in\mathbb{R}_+^{K}$ denotes the weight vector
    whose elements are weights to penalize the false-negative cases for each class.
    $\bi{C}^{\text{fp}}\in\mathbb{R}_+^{K\times K}$ is a square matrix with a zero diagonal.
    The $(k,l)$th element of $\bi{C}^{\text{fp}}$, where $k\ne l$, represents the weight parameter for penalizing false positives when a sample of class $l$ is misclassified to class $k$.
    For our goal, we should set the $\kappa$th element of $\bi{c}^{\text{fn}}$ to a value greater than $1$ and set the other elements to $1$.
    We also set $\kappa$ columns and $\kappa$ rows of $\bi{C}^{\text{fp}}$, except for the $(\kappa, \kappa)$th element, to values greater than $1$, diagonal elements to $0$, and the other elements to $1$.

    \subsubsection{L2-constrained softmax loss}
    L2-constrained softmax loss\cite{ranjan2017l2}, which explicitly treats the angle between features and weights as a loss,
    is given by
    \begin{equation}
        \mathcal{L}_{\text{L2}}=-\frac{1}{N}\sum_{n=1}^{N}\log \left(\frac{\exp \left(
        s\cos \left( \theta_{n,t_n} \right) \right)}
        { \sum_{k=1}^{K}\exp  \left(
        s\cos \left( \theta_{n,k} \right)\right)}\right),
        \label{eq:l2}
    \end{equation}
    where $s\in \mathbb{R}_+$ denotes the inverse of the temperature parameter.
    $\theta_{n,k}$ is given by
    \begin{equation}
        \theta_{n,k}=\arccos\left(\boldsymbol{W}_{:,k}\ensuremath{^{'\text{T}}} \boldsymbol{z}'_n\right),
        \label{eq:121}
    \end{equation}
    where $\boldsymbol{W}_{:,k}\ensuremath{^{'}}$ and $\boldsymbol{z}'_n$ are normalized respectively so that each $L^2$-norm of $\boldsymbol{W}_{:,k}$ and $\boldsymbol{z}_n$ is $1$.
    $\theta_{n,t_n}$ is given by \eqref{eq:121} when $k=t_n$.
    Equation~\eqref{eq:121} means that $\theta_{n,k}$ is the radian angle between the weight vector $\boldsymbol{W}_{:,k}\ensuremath{^{'}}$ of the last FC layer and the feature vector $\boldsymbol{z}'_n$.
    Hence, minimizing \eqref{eq:l2} is to minimize the angle $\theta_{n, t_n}$ when $k=t_n$ in \eqref{eq:121}.

    \subsubsection{ArcFace}
    Finally, we describe ArcFace as a method that explicitly penalizes the angle between features and weights.
    Deng \textit{et al.}\cite{deng2019arcface} proposed ArcFace as follows:
    \mathindent=2mm
    \small{
    \begin{eqnarray}
        &&\mathcal{L}_{\text{ArcFace}}= \nonumber\\
        & &\!-\!\frac{1}{N}\! \sum_{n=1}^{N}\! \log \!\frac{\exp \!\left(s \cos \left(\theta_{n,t_n}\!+\!m\right)\right)}{\!\exp \!\left(s \cos\! \left(\theta_{n,t_n}\!+\!m\right)\right)\!+\!\sum\limits_{k \neq t_n}\! \exp \!\left(s \cos \theta_{n,k}\right)},
        \label{eq:12}
    \end{eqnarray}}
    \mathindent=7mm
    where $m \in \mathbb{R}_{+}$ denotes the margin penalty, $s$ denotes the inverse of the temperature parameter, and $\theta_{n,k}$ is given by \eqref{eq:121}.
    ArcFace trains the model by adding an angular margin $m$ to $\theta_{n,t_n}$.
    The effect of adding the margin improves the separability between classes, which will be explained in detail in Section~\ref{34}.

    \section{Analysis of Class-Sensitive Separation}
    \label{sec:analysis}
    \begin{figure}[tb]
        \centerline{
            \includegraphics[scale=0.88]{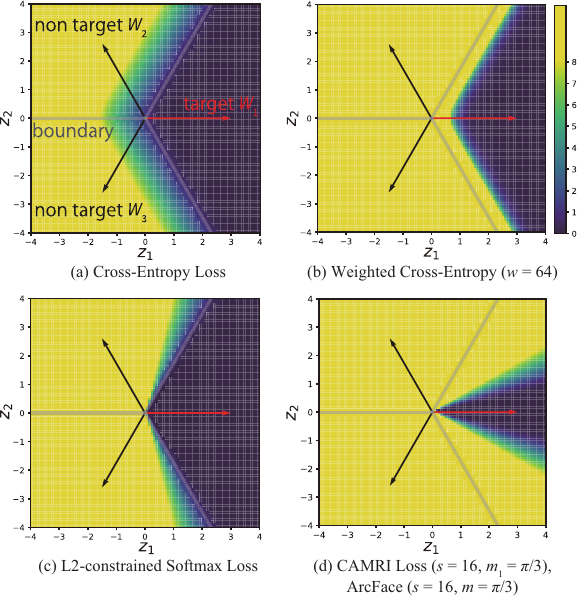}
        }
        \caption{
            The contour plots of the loss values formed by each loss function: (a), (b), (c), and (d) CE loss, WCE loss, L2-constrained softmax loss, and ArcFace or CAMRI loss, respectively.
            The vertical and horizontal axes represent the elements $z_1, z_2$ of \bi{z}, respectively, and the contour plots represent the loss values, with yellow and blue indicating high and low loss values, respectively.
            $\bi{W}_1, \bi{W}_2$, and $\bi{W}_3$ are the weight vectors of the last FC layer's weight.
            The ground-truth (GT) weight vector $\bi{W}_1$ is shown in the red arrow.
            The non-GT weight vectors $\bi{W}_2$ and $\bi{W}_3$ are shown in black arrows.
            The gray lines are the decision boundary.
        }
        \label{fig:2}
    \end{figure}

    For the problem described in Section~\ref{sec:introduction},
    we need to improve the recall of an important class without sacrificing overall accuracy.
    As mentioned in Subsection~\ref{sec:related}, threshold tuning is not helpful.
    To achieve this goal, we need to find a feature representation such that the feature vectors of the important class are well-separated from feature vectors in the other classes.
    In this section, we investigate the property of existing loss functions in the feature space and how the separation in the feature space is induced when we penalize the important class using the existing loss functions.

    \subsection{Setting}
    To visualize the properties of existing loss functions in detail, we consider a three-class classification problem, with the dimension of the feature vector is set to two.
    Fig.~\ref{fig:2} depicts the contour plots of loss values in the two-dimensional (2D) feature space for each loss function.
    Each loss function corresponding to these figures is CE loss (Fig.~\ref{fig:2}(a)), WCE loss(Fig.~\ref{fig:2}(b)), L2-constrained softmax loss (Fig.~\ref{fig:2}(c)), and ArcFace (Fig.~\ref{fig:2}(d))\footnote{This is consistent with the CAMRI loss defined later.}.
    We give a detailed explanation of Fig.~\ref{fig:2} when using the CE loss as an example (Fig.~\ref{fig:2}(a)).
    Because this is a three-class classification, there are three corresponding weight vectors in the last FC layer $\bi{W}_1,\bi{W}_2$, and $\bi{W}_3$.
    Fig.~\ref{fig:2}(a) depicts a contour plot of the CE loss in the 2D feature space when the GT label is $1$, which we set as the important class.
    In Fig.~\ref{fig:2}, the weight vector $\bi{W}_1$ corresponding to class $1$, which we call the GT weight vector, is shown in the red arrow.
    The weight vectors $\bi{W}_2$ and $\bi{W}_3$ corresponding to classes $2$ and $3$ are shown in black arrows.
    The blue area in the direction of $\bi{W}_1$ indicates that the loss values are low (because the GT label is $1$), and the yellow area in the opposite direction indicates high loss values.

    The last FC layer evaluates the inner product of the weight vector $\bi{W}_1$
    and the feature vector \bi{z} as follows:
    \begin{equation}
        \bi{W}_1\T \bi{z}= \| \bi{W}_1\| \|\bi{z} \| \cos \left( \theta_1 \right),
        \label{eq:110}
    \end{equation}
    where $\theta_1$ denotes the angle between $\bi{W}_1$ and $\bi{z}$, and a bias term shall not be considered.
    With the representation of \eqref{eq:110}, the CE loss, \eqref{eq:1}, is given by
    \mathindent=0mm
    \begin{eqnarray}
        \mathcal{L}_{\text{ce}}=
         \!-\frac{1}{N}\sum_{n=1}^{N}\log \left(\frac{\exp \left(
        \| \bi{W}_{t_n}\| \|\bi{z}_n \|\cos \left( \theta_{n,t_n} \right) \right)}
        { \sum_{k=1}^{K}\exp  \left(
        \| \bi{W}_{k}\| \|\bi{z}_n \|\cos \left( \theta_{n,k} \right)\right)}\right).
        \label{eq:010}
    \end{eqnarray}
    \mathindent=7mm
    Equation~\eqref{eq:010} can be minimized by minimizing $\theta_{t_n}$, which is realized by decreasing the angle between the features and GT weight vector.
    Therefore, it is important for high accuracy to train the model to represent \bi{z} such that $\theta_{t_n}$ is minimized.

    In the contour plot of CE loss (Fig.~\ref{fig:2}(a)), the region where the loss has low values spreads in the direction of $\bi{W}_1$.
    If we could sharpen the angle of the valley-like landscape represented by blue in the contour plot,
    we can expect that the separation between classes in the feature space is increased, yielding higher recall and accuracy.

    \subsection{Comparison between with and without weighting penalty for CE loss}
    \label{32}
    Next, we compared CE loss (Fig.~\ref{fig:2}(a)) and WCE loss(Fig.~\ref{fig:2}(b)), which is a CE loss penalizing the important class.
    We omit the discussion comparing with CRWWCE loss because the contour plots of the loss value obtained using CRWWCE loss have shapes similar to those of WCE loss (see Fig.~\ref{fig:a} in \ref{ap:omitplot}).
    Even with penalization to the important class, we can see that the angle discussed earlier is not sharpened.
    Instead, the counter is shifted toward the direction of $\bi{W}_1$.
    Because WCE loss simply increases the loss values for the important class by multiplying a constant, it is expected that the recall of an important class improves.
    However, because WCE loss does not necessarily reduce the angle between the features of the important class and the corresponding weight vector,
    higher accuracy than CE loss cannot be expected.
    In addition, because of the limited ability of the model to represent features, it is difficult to both maintain accuracy and improve the recall of an important class.
    Actually, as shown in Fig.~\ref{fig:5}, WCE loss improves the recall while sacrificing accuracy.
    Therefore, the cost-sensitive learning approach, including WCE loss and CRWWCE loss, would not work well for maintaining accuracy.

    \subsection{Comparison of CE loss and L2-constrained softmax loss}
    We now focus on Fig.~\ref{fig:2}(c).
    L2-constrained softmax loss\cite{ranjan2017l2}, defined by~\eqref{eq:l2}, explicitly treats the angle between features and weights as a loss.
    Comparing Figs.~\ref{fig:2}(a) and \ref{fig:2}(c), the shapes of the contours differ, whereas the angular sharpness of the valley-like landscape is the same.
    Therefore, we cannot expect to improve the recall of a specific class because this approach does not improve the separation between classes.

    \subsection{Comparison of angular-based loss with and without margin}
    \label{34}
    ArcFace, defined by~\eqref{eq:12}, introduces additive constant penalty $m$ to $\theta_{n, t_n}$ of L2-constrained softmax loss, which is called additive angular margin.
    Comparing the case without (Fig.~\ref{fig:2}(c)) and with (Fig.~\ref{fig:2}(d)) margin penalty, we can see that the angular sharpness of the valley-like landscape is sharpened.
    Because the angular margin corresponds to the geodesic distance margin on a hypersphere\cite{deng2019arcface}, ArcFace would improve the separation by making the feature distribution on the hypersphere compact, called intraclass compactness.
    Therefore, the margin penalty tuning is expected to improve the generalization ability of the classifier.
    However, ArcFace is unsuitable for improving the recall of an important class because ArcFace, which improves the separability of all classes, cannot concentrate on improving the separability of a specific class.

    \subsection{Proposal of CAMRI Loss}
    Considering the discussion in the last section, we expect that improvement in the recall of a specified important class without sacrificing overall accuracy is attained by
    (1) concentrating feature vectors of the important class along with the GT weight vector and
    (2) separating feature vectors from non-GT weight vectors.

    For this purpose, we propose CAMRI loss, which adds a margin only to the important class.
    Let $\kappa$ be an important class.
    Using a one-hot vector and $\mu \in \mathbb{R}_+$, and the class-sensitive angular margin vector is given by
    \begin{equation}
        \bi{m} = \mu[\hdots,0,1, 0, \hdots]\T,
        \label{eq:m}
    \end{equation}
    where only the $\kappa$th element is $\mu\geq0$ and the other elements are $0$.
    Using \eqref{eq:m}, CAMRI loss is defined by
    \mathindent=-2mm
    \small{
        \begin{eqnarray}
            &&\mathcal{L}_{\text{CAMRI}}= \nonumber\\
            &&\!-\!\frac{1}{N}\! \sum_{n=1}^{N}\! \log \!\frac{\exp \!\left(s \cos \left(\theta_{n,t_n}\!+\!m_{t_n}\right)\right)}{\!\exp \!\left(s \cos\! \left(\theta_{n,t_n}\!+\!m_{t_n}\right)\right)\!+\!\sum\limits_{k \neq t_n}\! \exp \!\left(s \cos \theta_{n,k}\right)},
            \label{eq:13}
        \end{eqnarray}}
%
    \mathindent=7mm
    where $\theta_{n,k}$ is given by~\eqref{eq:121} and $s$ denotes the inverse of the temperature parameter.
    For each training sample, if the label $t_n$ is $\kappa$, the class-sensitive angular margin $m_{t_n} = \mu$ is added; otherwise, no margin is added.
    Notably, the class-sensitive angular margin is applied only in the training process.

    Because CAMRI loss adds the angular margin only to the important class, the intraclass compactness of the important class is expected to increase compared with the other classes.

    \section{Experiments}
    \label{sec:experiment}
    \begin{figure*}[tb]
        \centerline{
            \includegraphics[width=175mm]{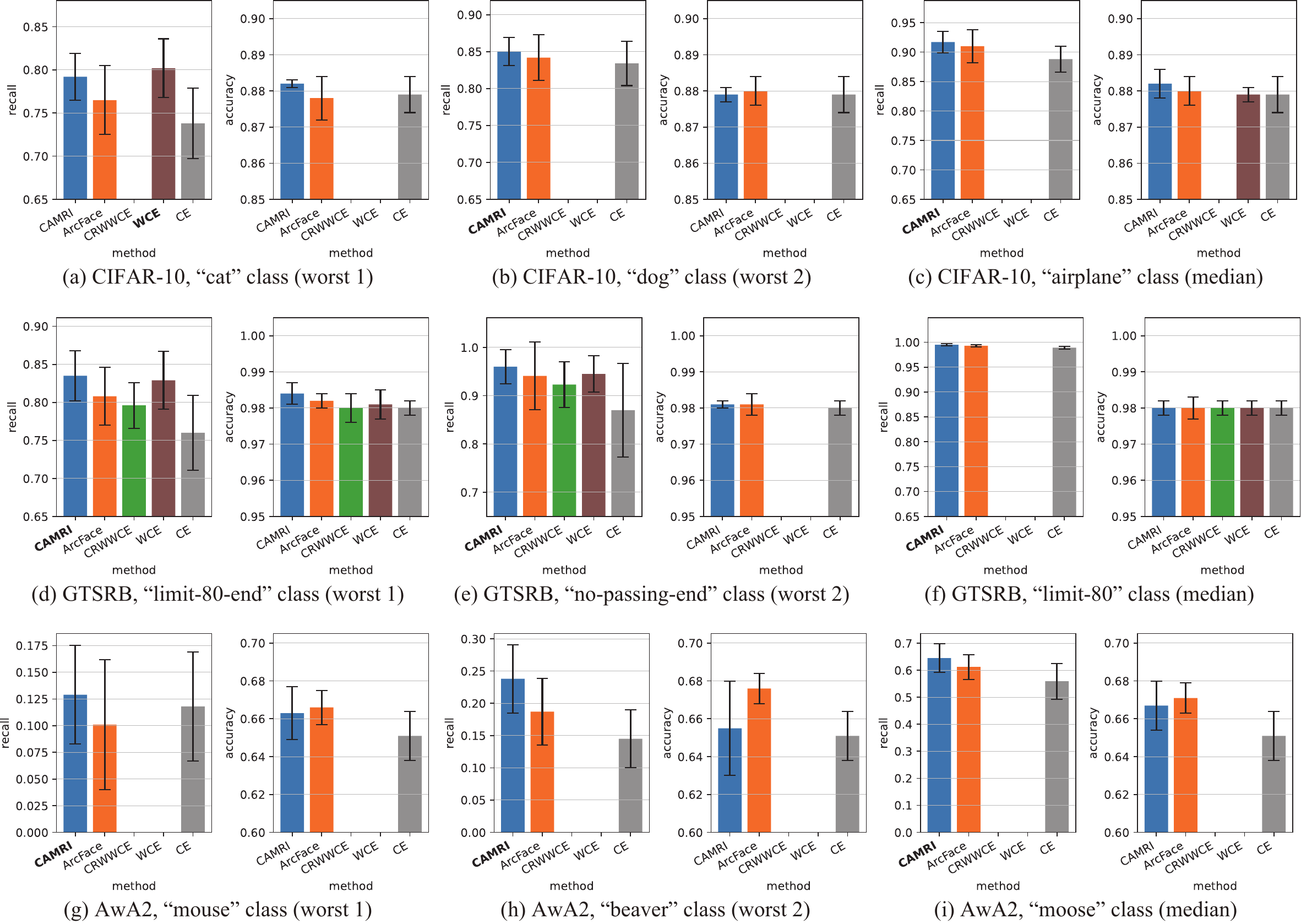}
        }
        \caption{
            The mean and standard deviation of the recall of the important class and accuracy over ten trials, obtained from the criteria described in Section~\ref{sec:criteria}, are shown for each important class setting,
            as indicated in their subcaptions, for example, (a) indicates the result when the cat class, which is the worst 1 recall when learning CIFAR-10 with CE loss, is set as the important class.
            Settings (b) through (i) also correspond to their subcaptions.
            In each setting from (a) to (i), the left figures represent recall and accuracy, respectively.
            The horizontal axes are the compared methods, and the vertical axes are the recall or accuracy values.
            The error bars mean standard deviation.
        }
        \label{tab:1}
    \end{figure*}

    \subsection{Experiment Setting}

    \subsubsection{Datasets}
    We employed three datasets, CIFAR-10\cite{krizhevsky2009learning}, GTSRB\cite{Stallkamp-IJCNN-2011},
    and AwA2\cite{8413121}, for the multiclass classification problem.
    CIFAR-10 is a 10-class dataset with a balanced amount of data per class and has 50,000 training images and 10,000 test images of size $32\times32\times3$.
    GTSRB is a color image dataset of German road signs;
    it has 43 classes with an imbalanced amount of data per class and contains 39,209 training images and 12,630 test images.
    We resized them to $48\times48\times3$ because the image sizes vary from $15\times15\times3$ to $250\times250\times3$.
    AwA2 has 50 classes with an imbalanced amount of data per class and contains 37,322 animal images.
    We resized the image sizes to $64\times64\times3$ and divided the dataset into 26,157 training images and 11,165 test images.
    In all datasets, we normalized all pixels so that the pixel values were within the range of $[0, 1]$.

    \subsubsection{CNN Setups}
    We used Tensorflow\cite{abadi2016tensorflow} to implement loss functions and the DCNN model.
    The DCNN model comprises a feature extractor and a linear classifier.
    The feature extractor contains convolution layers\cite{lecun1998gradient, krizhevsky2012imagenet}, batch normalization layers\cite{ioffe2015batch}, max-pooling layers\cite{lecun1998gradient}, dropout layers\cite{srivastava2014dropout}, and a global average pooling (GAP) layer\cite{lin2013network}.
    The linear classifier contains two FC layers.
    ReLU function\cite{nair2010rectified} activates the FC and convolution layers other than the last FC layer.

    During training, the last FC layer is not activated for ArcFace and CAMRI loss; it is activated by the softmax function in other methods.
    During the evaluation, the last FC layer is activated by the softmax function, and accuracy and recall are obtained.
    Details of the model structure are shown in \ref{ap:model}.

    We set the batch size to $64$ and the number of epochs to $300$.
    We used Adam\cite{kingma2014adam} with the a $0.001$ learning rate for optimization.

    \subsubsection{Hyperparameter Settings}
    \label{sec:param}
    The ranges of weight and margin parameters for each loss function were set as follows by preliminary experiments so that the recall of the important class becomes high.
    In the following, let $\kappa$ be the index of an important class.

            {\bf WCE.}
    The weight vector \bi{w} of WCE loss is a $K$-dimensional vector.
    Its $\kappa$th element was set to $4, 8, 12,\cdots, 40$, and the other elements in \bi{w} were set to $1$.

            {\bf CRWWCE.}
    The weight vector $\bi{w}^{\text{fn}}$ and matrix $\bi{W}^{\text{fp}}$ of CRWWCE loss were set as follows.
    $\bi{w}^{\text{fn}}$ is a $K$-dimensional vector.
    Its $\kappa$th element was set to $1, 4, 8, 12,\cdots, 40$, and the other elements in $\bi{w}^{\text{fn}}$ were set to $1$.
    $\bi{W}^{\text{fp}}$ is a $K\times K$ matrix, and each element was set as follows:
    (1) all elements in the $\kappa$th row and $\kappa$the column were varied as $1.0,1.2, \cdots, 4.0$, (2) all diagonal elements were set to $0$,  and (3) the other elements were set to $1$.

            {\bf ArcFace.}
    The scalar margin parameter $m$ was set to $0,\frac{\pi}{64},\frac{2}{64}\pi,\cdots,\frac{8}{64}\pi$.
    The inverse of the temperature parameter $s$ was set to $2^i$, where $i$ was varied as $i=0,1,\cdots,6$.

            {\bf CAMRI loss.}
    The class-sensitive angular margin vector \bi{m} of CAMRI loss is a $K$-dimensional vector.
    Its $\kappa$th element was set to $0,\frac{\pi}{64},\frac{2}{64}\pi,\cdots,\frac{8}{64}\pi$ and the other elements were set to $0$.
    The inverse of the temperature parameter $s$ was set to $2^i$, where $i$ was varied as $i=0,1,\cdots,6$.

    \subsection{Performance Evaluation}
    \label{sec:eval}

    \subsubsection{Evaluation Methods}
    \label{sec:criteria}
    We compared the recall of each important class and the accuracy of models obtained by training with CAMRI loss, CE loss, ArcFace, CRWWCE loss, and WCE loss.
    Then, we investigated whether the recall of an important class is improved without sacrificing accuracy.

    We tested our models with three important classes with the worst, second-worst, and median recall among all classes when the model had been trained with CE loss.
    With this criterion, \{cat, dog, airplane\}, \{limit-80-end, no-parking-end, limit-80\}, and \{mouse, beaver, smooth\},
    were selected as the important classes respectively for CIFAR10, GTSRB, and AwA2, respectively.

    We performed ten training trials with all combinations of the parameters described in Subsection~\ref{sec:param}.
    The average recall and accuracy of the ten trials are compared and evaluated using the following criteria:
    \begin{enumerate}
        \item Recall and accuracy obtained by using the CE loss are set as the baseline.
        \item Among the hyperparameter settings specified in Subsection~\ref{sec:param}, the results of methods that maintain equal or better accuracy compared with the baseline are selected.
        The results of methods that do not achieve equal accuracy in any parameter setting are not selected.\label{itm:ext}
        \item The results obtained using the parameter setting that achieves the highest recall for each method are shown in Figs.~\ref{tab:1}(a)-\ref{tab:1}(i).
        \item For each important class setting, the method that achieves the highest recall is shown in bold letters.
    \end{enumerate}

    \subsubsection{Results}
    The results are summarized in Fig.~\ref{tab:1}.
    The hyperparameters employed to obtain Fig.~\ref{tab:1} are summarized in Table~\ref{tab:a2} in \ref{ap:hypara}.

    Figs.~\ref{tab:1}(a)-\ref{tab:1}(i) correspond to each critical class setting;
    in each setting, the left and right figures denote recall and accuracy, respectively.
    As mentioned above, improving recall with scarifying accuracy is extremely easy.
    For this reason, some data in Fig.~\ref{tab:1} are left blank when the accuracy is less than the baseline (CE loss), even if the corresponding recall is improved.
    For reference, Table~\ref{tab:aa1} in \ref{ap:onlyrecall} shows the methods that attain the maximum recall, where recalls are displayed even when the accuracies are lower than the baselines.

    From Fig.~\ref{tab:1}, the proposed method, CAMRI loss, improved the recall of the important classes in all settings.
    In addition, ArcFace improved the recall of the important classes in all settings except for the mouse class of AwA2.
    Comparing CAMRI loss and ArcFace, CAMRI loss achieved higher recalls than ArcFace in all cases,
    suggesting that adding angular margin only to the important class induces better recalls.
    By contrast, CRWWCE and WCE losses cannot improve the recall by maintaining the accuracy in several cases.
    Although CRWWCE and WCE losses improved recall, the degree of improvement was not as great as that of CAMRI loss, except for WCE loss in Fig.~\ref{tab:1}(a).

    As described in Section~\ref{sec:analysis}, the existing methods do not sharpen the angle
    between the distribution of feature vectors and GT weight vectors and cannot improve the recall while maintaining accuracy.
    In fact, CRWWCE and WCE losses can only improve recall by sacrificing accuracy (See Table~\ref{tab:aa1} in \ref{ap:onlyrecall} for details).

    Therefore, CAMRI loss, which penalizes the angle between the features of the important class and the corresponding weight vector, can achieve higher recall than existing methods that penalize CE loss by weighting corresponding to the important class without sacrificing accuracy.

    The reason for the low recall of the mouse and beaver classes in AwA2 is that these classes contain more examples that are unclear and difficult to learn than the classes with higher recall.
    Please refer to Subsection \ref{sec:awa} for details.

    \subsection{Effects of margin on separability}
    \begin{figure}[tb]
        \centerline{
            \includegraphics[width=85mm]{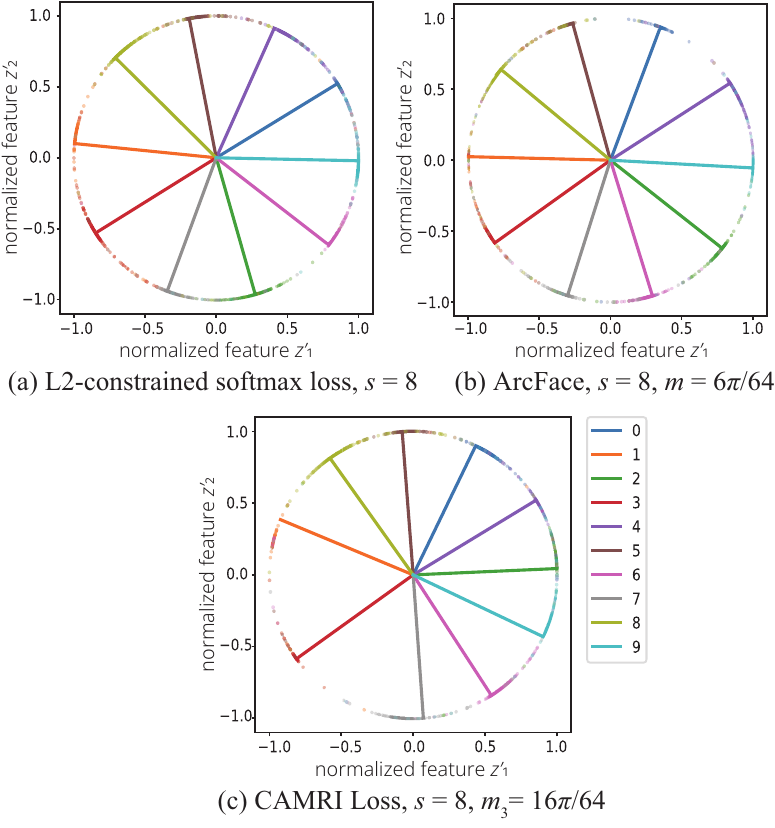}
        }
        \caption{
            Normalized feature vectors of MNIST $\bi{z}^{\prime}$ (represented as dots) and weight vectors (represented as solid lines) in the two-dimensional feature space:
            (a), (b), and (c) L2-constrained softmax loss, ArcFace, and CAMRI loss, respectively.
            The vertical and horizontal axes represent the elements $z_1^{\prime}, z_2^{\prime}$ of $\bi{z}^{\prime}$, respectively.
            In the results, class "3" (shown in red) was set as the important class.
        }
        \label{fig:6}
    \end{figure}
    From the results in Fig.~\ref{tab:1}, we can see that adding the angular margin only to the important class (i.e., CAMRI loss) improves the recall more than adding the angular margin to all classes equally (i.e., ArcFace).
    We investigated the reason for this result by visualizing the feature space and observing the effect of the class-sensitive angular margin on the separability of features.

    For the visualization purpose, we trained a CNN with a 2D feature space using MNIST\cite{lecun1998gradient}.
    Fig.~\ref{fig:6} shows the GT weight vector and the corresponding distribution of normalized feature vectors in the feature space when the CNN is trained with loss function without angular margin (L2-constrained softmax loss),
    with the angular margin (ArcFace), and with the class-sensitive angular margin (CAMRI loss).
    We set "3" as the important class (shown in red).

    Comparing the case without margin (Fig.~\ref{fig:6}(a)) and with an equal margin for all classes (Fig.~\ref{fig:6}(b)),
    there is no improvement in the separability of class "3" (represented by red).
    Moreover, when the margin is added only to "3" (Fig.~\ref{fig:6}(c)), the angle between the weight vector for "3" and the adjacent weight vectors are well-separated.
    The features are distributed in the direction of the weight vector for "3," and the separability of class "3" from other classes is relatively increased.
    This increase in the important class separability is caused by adding the angular margin penalty only to the important class, yielding an improvement in the recall.

    \begin{table}[tb]
        \caption{
            The standard deviation of the radian angle between the feature vectors and the corresponding GT weight vector for each method is shown.
            The results are cases when each of the cat, dog, and airplane classes of CIFAR-10 are set to the important class.
            In each case, the standard deviation of the important class, the smallest standard deviation among those excluding the important class, and the median standard deviation among those excluding the important class are displayed.
        }
        \label{tab:2}
        \begin{center}
            \footnotesize
            \scalebox{0.855}{
                \begin{tabularx}{1.15\linewidth}{p{9.27em}CCCCC}
                    \toprule
                    \multicell{}{class}   & CAMRI           & ArcFace         & \multicell{WCE}{($w=4$)} & \multicell{ArcFace}{($m=0$)} & CE \\ \midrule
                    cat (important)       & \textbf{0.0695} & 0.0384          & 0.0603                   & \textbf{0.0928}              & 0.0842          \\
                    min excl. cat         & 0.0948          & \textbf{0.0168} & \textbf{0.0540}          & 0.0937                       & \textbf{0.0570} \\
                    median excl. cat      & 0.1143          & 0.0403          & 0.0766                   & 0.1009                       & 0.0806          \\  \midrule
                    dog (important)       & \textbf{0.0717} & 0.0256          & 0.0667                   & 0.1009                       & 0.0815          \\
                    min excl. dog         & 0.0919          & \textbf{0.0158} & \textbf{0.0602}          & \textbf{0.0928}              & \textbf{0.0570} \\
                    median excl. dog      & 0.1116          & 0.0328          & 0.0835                   & 0.1009                       & 0.0806          \\  \midrule
                    airplane (important)  & \textbf{0.0367} & 0.0512          & 0.0727                   & 0.0512                       & 0.0806          \\
                    min excl. airplane    & 0.0439          & \textbf{0.0441} & \textbf{0.0627}          & \textbf{0.0441}              & \textbf{0.0570} \\
                    median excl. airplane & 0.0558          & 0.0555          & 0.0714                   & 0.0555                       & 0.0815          \\ \bottomrule
                \end{tabularx}
            }
        \end{center}
    \end{table}

    \subsection{Reducing Intraclass angular variance of important class by a class-sensitive angular margin}
    In this subsection, we observed that adding the class-sensitive angular margin enhances the intraclass compactness of feature vectors in the feature space.
    To measure the intraclass compactness, we measured the standard deviations of the angle between the GT weight vectors and corresponding feature vectors.

    Table~\ref{tab:2} shows the standard deviations of the angles when the cat, dog, and airplane classes are set as the important class (average of ten trials).
    For each important class, the minimum and the median excluding the important class are also shown.
    Bold texts indicate the smaller value of the important class and other minimum.
    The parameter values of CAMRI loss and ArcFace follow the setting in Fig.~\ref{tab:1}.
    WCE loss is measured with a fixed value of $w=4$.
    The value of $s$ for ArcFace ($m=0$) is the same as that of CAMRI loss.

    Comparing the standard deviation of the important class and the other classes, CAMRI loss makes the standard deviation of the important class $0.783$ times less than that of the minimum excluding of the important class.
    By contrast, the other methods do not necessarily make the standard deviation of the important class lowest,
    meaning that they do not improve the intraclass compactness even with penalization.

    Notably, our method does not necessarily attain the lowest standard deviation for the important class among other methods.
    The standard deviation of the important class is made lower than that of other classes relatively,
    which is considered sufficient to improve the recall of the important class.

    \subsection{Effects of penalty changes on recall and accuracy}
    \begin{figure}[tb]
        \centerline{
            \includegraphics[width=80mm]{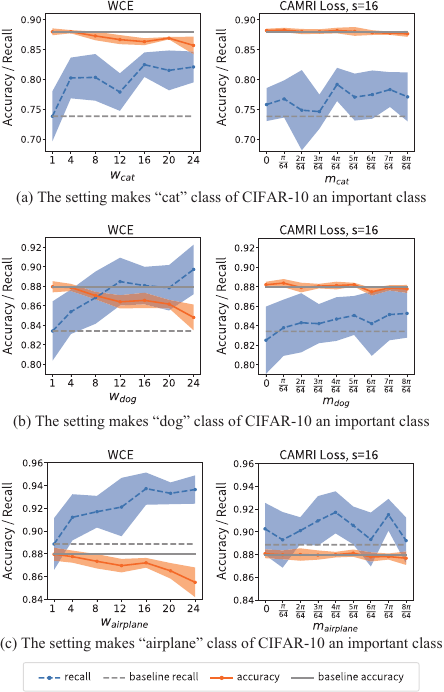}
        }
        \caption{
            Changes in recall of the important class (blue line) and accuracy (orange line) by varying penalties are shown.
            The lines show the mean value in ten trials, and the bands show the standard deviation.
            The left columns are WCE loss, and the right columns are CAMRI loss.
            (a)-(c) are the cat, dog, and airplane classes of CIFAR-10, respectively, as the important class.
            The horizontal axis is the value of the penalty ($w_\kappa$ for WCE loss and $m_\kappa$ for CAMRI loss), whereas the vertical axis is the value of recall of the important class and accuracy.
        }
        \label{fig:5}
    \end{figure}
    \begin{figure}[tb]
        \centerline{
            \includegraphics[width=70mm]{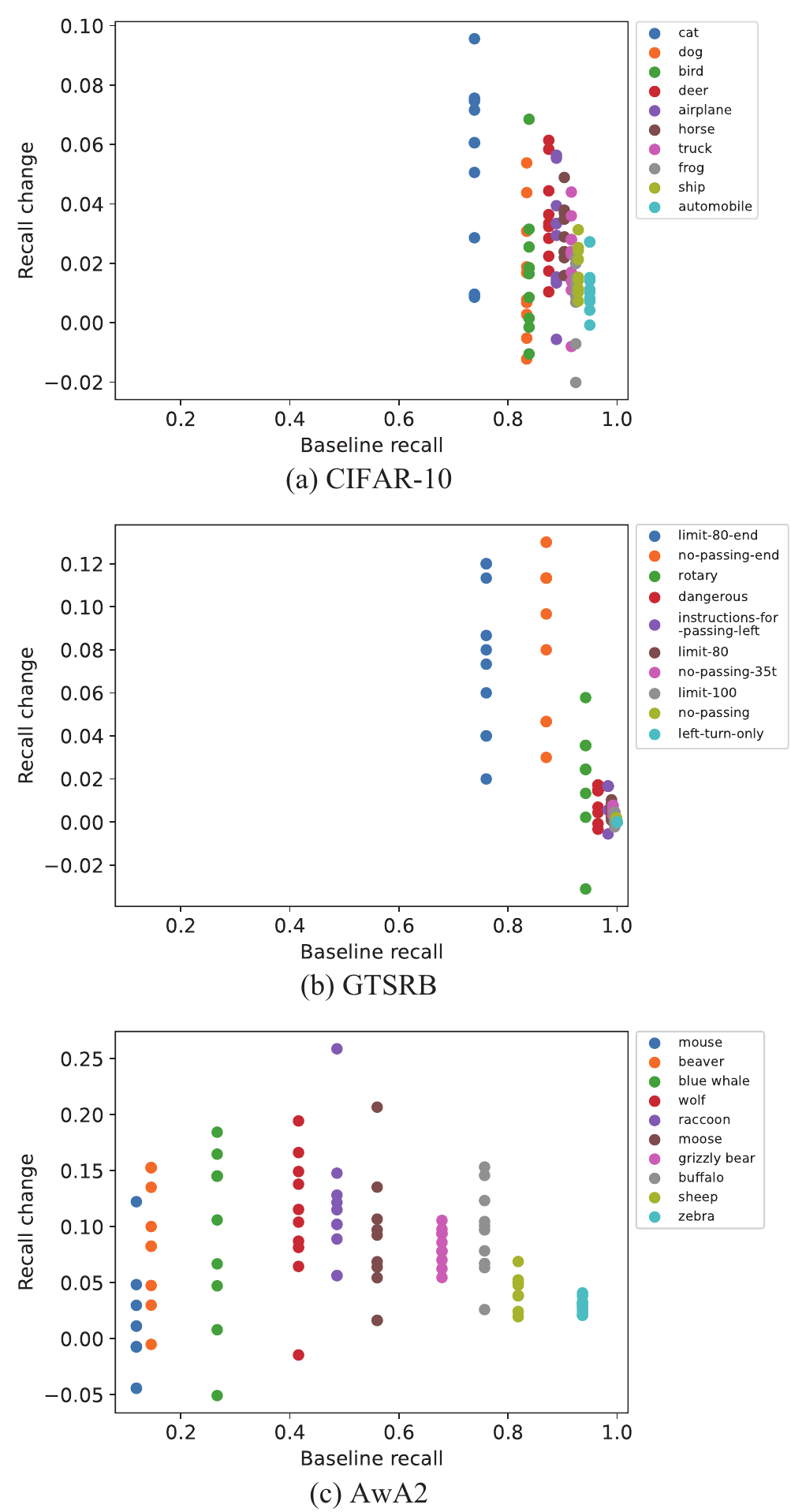}
        }
        \caption{
            Comparison of recall improved by CAMRI loss on (a)-(c) CIFAR-10, GTSRB, and AwA2, respectively.
            The horizontal axis shows baseline recalls obtained by CE loss, averaged over ten trials.
            The vertical axis shows changes in recall when trained with CAMRI loss for ten trials.
            The color of the dots corresponds to the class, and the results of ten trials are shown in the figure.
        }
        \label{fig:ad1}
    \end{figure}
    \begin{figure*}[tb]
        \centerline{
            \includegraphics[width=181mm]{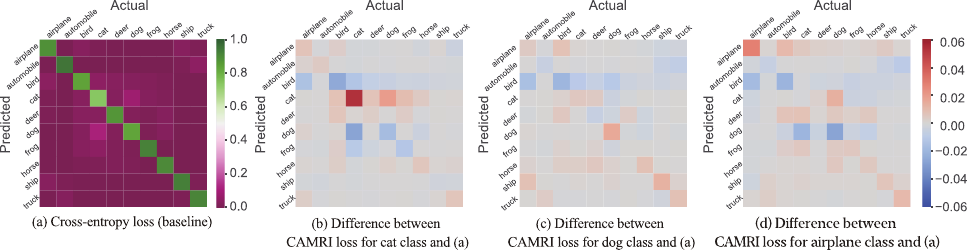}
        }
        \caption{
            The left (a) is the baseline confusion matrix of CIFAR-10 obtained from trained using CE loss.
            (b), (c), and (d) Differences between the confusion matrix with CE loss (i.e., (a)) and that with the CAMRI loss, where the cat, dog, and airplane classes are set as the important class, respectively.
            Each result is the average of ten trials and corresponds to the result shown in Fig.~\ref{tab:1}.
        }
        \label{fig:3}
    \end{figure*}

    For further investigation of the relationship between the penalty given to the important class and recall/accuracy, we evaluated the changes in recall and accuracy with respect to penalty parameters of CAMRI loss and WCE loss.
    The cat, dog, and airplane classes of CIFAR-10 are trained as the important class by changing the penalty parameters.

    Fig.~\ref{fig:5} shows how the recall and accuracy change when the class-sensitive angular margin $m_\kappa$ of CAMRI loss and the weight penalty $w_\kappa$ of WCE loss change,
    where $\kappa$ is the index of the important class.
    As $w_\kappa$ increases, WCE loss improves the recall, but on the contrary, sacrifices the accuracy.
    Meanwhile, as $m_\kappa$ increases, CAMRI loss improves the recall while maintaining accuracy sufficiently.

    The results show that CAMRI loss improves the recall by acquiring a feature representation having better intraclass compactness without sacrificing the accuracy,
    whereas WCE loss improves the recall by controlling the tradeoff of recall and accuracy.

    \subsection{Possibility of improving recall of any class}

    We examined under what conditions a recall can be significantly improved.
    We selected ten classes of CIFAR-10, GTSRB, and AwA2, and set each class as important in each case.
    For CIFAR-10, all classes were selected.
    For GTSRB and AwA2, nine classes located at six intervals in the order of lowest baseline recall and one class with the second lowest recall were selected.
    We trained models with CAMRI loss with the same parameter setting range as in Subsection \ref{sec:param}, and evaluated the improvement in recall with the same criterion as in Subsection \ref{sec:criteria}.

    In Fig.~\ref{fig:ad1}, the horizontal axis shows baseline recalls obtained by CE loss, average over ten trials.
    The vertical axis shows changes in recall when trained with CAMRI loss for ten trials.
    Because each data point is sampled according to the criteria described in Subsection \ref{sec:criteria}, the average value of data points maintains accuracy compared with CE loss in each class.

    Fig.~\ref{fig:ad1} suggests that no matter which class of CIFAR-10 is set as the important class, CAMRI loss can improve the recall without sacrificing accuracy in many cases.
    In addition, the lower the baseline recall for a class, the greater the recall improvement for that class, but the recall improvement tends to be lower when the baseline recall is lower than a certain level.
    For example, in Fig.~\ref{fig:ad1}(a), the change in recall for the worst baseline recall in CIFAR-10 (blue) attained greater improvement in recall than other classes having higher baseline recall.
    Meanwhile, in GTSRB and AwA2, the classes with the most improved recall among the ten classes are the second (orange in Fig.~\ref{fig:ad1}(b)) and nineteenth (purple in Fig.~\ref{fig:ad1}(c)) in order of lowest baseline recall, respectively, and recall improvements of classes with lower baseline recall than those classes are lower as the baseline is lower.
    In the classes with lower baseline recall than those with the best recall improvement among the ten classes, the lower the baseline recall, the lower the recall change.
    In Fig.~\ref{fig:ad1}(c), classes with lower changes in recall as baseline recall decreases are considered to be more difficult for the model to learn than classes that are not.
    Thus, it is expected that recall improvement with CAMRI loss can improve recall for any class depending on the difficulty of learning class features and baseline recall when learning with CE loss.

    \subsection{Impacts of improving recall of the important class on the other classes}
    When CAMRI loss improves the recall of an important class while maintaining accuracy, it is thought that the recall of other classes may be decreased.
    Therefore, we observed the difference in confusion matrices with CE loss and CAMRI loss.

    Fig.~\ref{fig:3}(a) shows the confusion matrix when CIFAR-10 is trained with CE loss.
    Fig.~\ref{fig:3}(b), (c), and (d) show the difference in the confusion matrix when CIFAR-10 is trained with CE loss and CAMRI loss,
    where the cat, dog, and airplane classes are set as the important classes, respectively.
    The average of ten training trials are shown.

    When the recall of the important class (say, class A) improves, some misclassified samples as the important class are correctly classified.
    In addition, the recall of some other classes (say, class B) decreases.
    Between classes A and B, there appears to be a trend of decreasing the number of class A samples misclassified as class B.
    Looking at Fig.~\ref{fig:3}(b), the recall of the cat class (important class) improves, whereas the recall of bird and dog classes decreases;
    also, the number of samples of the cat class misclassified as bird and cat misclassified as dog decreases, and the recall of the bird and dog classes decreases.
    Fig.~\ref{fig:3}(c) also shows the same tendency.
    The number of samples of the dog class (important class) misclassified as bird decreases, and the recall of the dog class improves whereas that of the bird class decreases.
    Similarly, Fig.~\ref{fig:3}(d) shows the number of samples of the airplane class (important class) misclassified as bird decreases,
    and the recall of the airplane class improves whereas the recall of the bird class decreases.

    These results show, first, that samples that were misclassified as the important class are correctly classified, contributing to improving recall of the important class.
    Second, the recall of classes to which samples that contribute to the recall improvement belong tends to decrease.
    These two factors suggest that when CAMRI loss improves the recall of the important class, the separation between the important class and another class is influenced.

    \subsection{Extension to multiple important classes settings}
    \begin{table*}[tb]
        \centering
        \caption{
            For CAMRI loss, the accuracy, recall of each important class, and $m_k$ obtained from the hyperparameter optimization are shown for each dataset and each multiple important class setting.
            For CE loss, the mean and standard deviation of the accuracy and recall of each important class over ten trials are shown on
            (a)-(c) CIFAR-10, GTSRB, and AwA2, respectively.
            $m_k$ means the class-sensitive angular margin of class $k$ of CAMRI loss for each important class.}
        \label{tab:ad2}
        \begin{tabular}{llccccc}
            \multicolumn{7}{c}{(a) CIFAR-10} \\
            \toprule
            & & \multicolumn{2}{c}{Two important class setting} & \multicolumn{3}{c}{Three important class setting} \\ \cmidrule(lr){3-4} \cmidrule(lr){5-7}
            method             & ~      & cat             & airplane        & cat             & deer            & truck           \\ \midrule
            ~                  & Recall & 0.802           & 0.936           & 0.815           & 0.911           & 0.939           \\
            CAMRI Loss         & $m_k$  & 0.274           & 0.170           & 0.234           & 0.160           & 0.025           \\
            ~ & Accuracy & \multicolumn{2}{c}{0.8823} & \multicolumn{3}{c}{0.8842} \\\midrule
            CE Loss & Recall & 0.738$\pm$0.041 & 0.888$\pm$0.022 & 0.738$\pm$0.041 & 0.875$\pm$0.023 & 0.916$\pm$0.021 \\
            ~ & Accuracy & \multicolumn{2}{c}{0.879$\pm$0.005} & \multicolumn{3}{c}{0.879$\pm$0.005} \\ \bottomrule
        \end{tabular}
        \begin{tabular}{llccccc}
            &        &                 &                 &                 &                 &                 \\\multicolumn{7}{c}{(b) GTSRB} \\
            \toprule
            &      & \multicolumn{2}{c}{Two important class setting}    & \multicolumn{3}{c}{Three important class setting}        \\ \cmidrule(lr){3-4} \cmidrule(lr){5-7}
            method                  & ~ & limit-80-end           & limit-80           & limit-80-end           & uneven-surface           & animal-crossing           \\ \midrule
            ~ & Recall & 0.846 & 0.995 & 0.933 & 0.991 & 1.000 \\
            CAMRI Loss & $m_k$ & 0.353 & 0.219 & 0.367 & 0.154     & 0.331      \\
            ~                  & Accuracy  & \multicolumn{2}{c}{0.981}           & \multicolumn{3}{c}{0.982}           \\\midrule
            CE Loss & Recall & 0.760$\pm$0.049 & 0.989$\pm$0.003 & 0.760$\pm$0.049 & 0.983$\pm$0.027 & 0.995$\pm$0.004 \\
            ~ & Accuracy & \multicolumn{2}{c}{0.980$\pm$0.002} & \multicolumn{3}{c}{0.980$\pm$0.002} \\ \bottomrule
        \end{tabular}
        \begin{tabular}{llccccc}
            &        &                 &                 &                 &                 &                 \\
            \multicolumn{7}{c}{(c) AwA2} \\
            \toprule
            & & \multicolumn{2}{c}{Two important class setting} & \multicolumn{3}{c}{Three important class setting} \\ \cmidrule(lr){3-4} \cmidrule(lr){5-7}
            method             & ~      & mouse           & moose           & mouse           & hippopotamus    & horse           \\ \midrule
            ~                  & Recall & 0.222           & 0.771           & 0.166           & 0.666           & 0.847           \\
            CAMRI Loss                  & $m_k$  & 0.039           & 0.361           & 0.485           & 0.286           & 0.118           \\
            ~ & Accuracy & \multicolumn{2}{c}{0.652} & \multicolumn{3}{c}{0.651} \\\midrule
            CE Loss & Recall & 0.118$\pm$0.051 & 0.559$\pm$0.066 & 0.118$\pm$0.051 & 0.459$\pm$0.072   & 0.685$\pm$0.062 \\
            ~ & Accuracy & \multicolumn{2}{c}{0.651$\pm$0.013} & \multicolumn{3}{c}{0.651$\pm$0.013} \\ \bottomrule
        \end{tabular}
    \end{table*}

    Up to this point, we have assumed that the number of important classes is one.
    However, in the real world, we often need to improve the recalls of multiple important classes.
    In this section, we investigate whether the CAMRI loss can improve the recall of multiple important classes without sacrificing accuracy.

    We employed the CIFAR-10, GTSRB, and AwA2 to investigate two settings: two important classes and three important classes.
    In the two-class setting, we used the worst and the fifth-worst classes as the important classes in the baseline.
    In the three-class setting, we used the worst, the fourth-worst, and the seventh-worst classes as the important classes in the baseline.
    Let $T$ be the number of classes subject to recall improvements and
    $\bi{\tau} \in \mathbb{R}^T$ be a vector of indices of classes subject to recall improvements.
    %
    We defined the class-sensitive angular margin vector \bi{m} as a vector whose each $\tau_i$th element $m_{\tau_i}$, where $1 \leq i \leq T$, is $m_{\tau_i} \geq 0$.
    The elements of \bi{m} excluded $\tau_i$th elements were set to $0$.
    The objective of sampling each $m_i$ was defined as the following multiobjective optimization problem:
    \begin{eqnarray}
        \max_{  \bi{m} } &&\left( \text {Accuracy}(f_{\bi{m}}), \right. \nonumber \\
        && \mbox{\ } \left.  \text {Recall}( f_{\bi{m}})_{\tau_{1}}, \cdots, \text {Recall}( f_{\bi{m}})_{\tau_{T}}\right),
        \label{eq:moo}
    \end{eqnarray}
    where $f_{\bi{m}}$ denotes a multiclass classification model trained by CAMRI loss with $\bi{m}$, $\text {Accuracy}(f)$ means the test accuracy of a model $f$, and $\text {Recall}(f)_{\tau_i}$ means the test recall of class $\tau_i$ of a model $f$.
    Each $m_{\tau_i}$ was searched in the range of $0.0$-$0.5$, and NSGA-II algorithm\cite{996017} is used to solve \eqref{eq:moo}.
    The parameter $s$ was fixed at $s=16$ for CIFAR-10 and AwA2 and $s=64$ for GTSRB because these values had been the most frequent value that attained the best recall observed in each dataset in the results of the experiments in Subsection~\ref{sec:eval} (See Table \ref{tab:a2} in \ref{ap:hypara} for details).
    We implemented the hyperparameter optimization in Optuna\cite{akiba2019optuna} and set the number of trials for searching $m_{\tau_i}$ to 50.

    The evaluation was performed with the following criteria:
    \begin{enumerate}
        \item The mean of recall and accuracy with CE loss are used as the baseline.
        \item Among the trials, the results that maintain equal or better accuracy than the baseline are selected.
        \item For each important class setting, the case where the sum of recall improvements of each class is the largest in the cases where the recall of each important class all improved is shown in Table \ref{tab:ad2}.
    \end{enumerate}

    In each table of Tables~\ref{tab:ad2} from (a)-(c), the lower part shows the accuracy and recall of each class when trained with CE loss, and the upper part shows the accuracy, recall, and margin values when trained with CAMRI loss and following the above criteria.
    The results show that CAMRI loss can improve recall of multiple important classes while maintaining accuracy compared to CE loss in all cases.
    Therefore, CAMRI loss can improve each recall of multiple important classes without sacrificing accuracy.

    \subsection{Causes of low recall in AwA2}

    \begin{figure*}[tb]
        \centerline{
            \includegraphics[width=175mm]{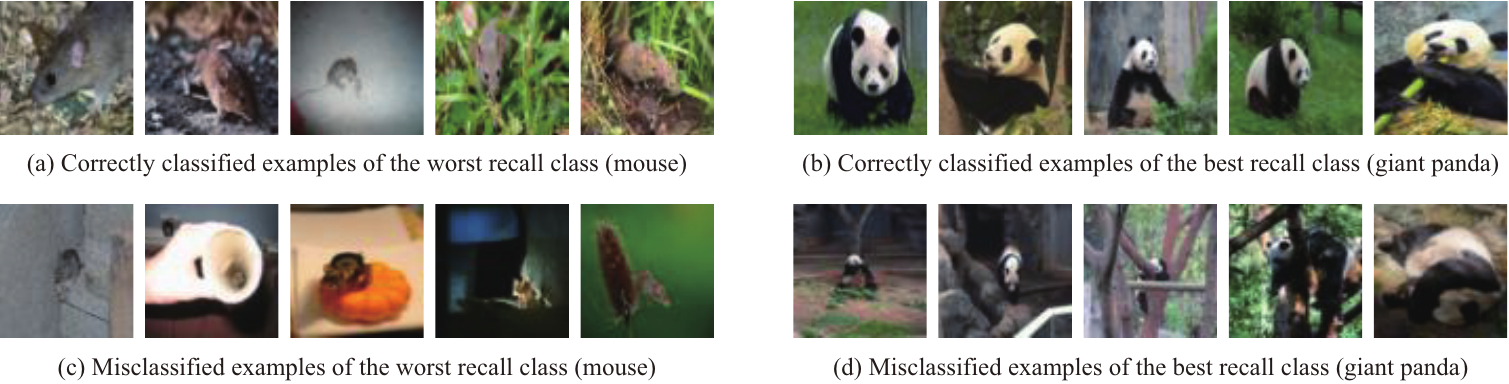}
        }
        \caption{
            Five random examples of correctly classified and misclassified examples are shown for each class of the worst and best recall when training AwA2 with CE loss, respectively.
            (a) and (b) represent correctly classified examples, and (c) and (d) represent misclassified examples. Also, (a) and (c) represent the worst recall class, and (b) and (d) represent the best recall class.
        }
        \label{fig:ad_awa}
    \end{figure*}

    \label{sec:awa}

    From Fig.~\ref{tab:1}, in AwA2, the baseline recalls of the mouse (worst 1) and beaver (worst 2) classes are about 0.118 and 0.145, respectively, which are significantly lower than in other datasets.
    Improving recall while maintaining accuracy with CAMRI loss improved recall by about 0.011 for the mouse class, but this improvement was insufficient.
    In this section, we discuss its causes.

    To compare well-classified classes and features and those that are not, some examples of the mouse class with the worst baseline recall and some examples of the giant panda class with the best baseline recall are shown in Fig.~\ref{fig:ad_awa}.
    Fig.~\ref{fig:ad_awa} is shown separately for correctly classified (Figs.~\ref{fig:ad_awa}(a),(b)) and misclassified (Figs.~\ref{fig:ad_awa}(c),(d)).
    The five examples shown in each case were selected at random.
    First, comparing Figs.~\ref{fig:ad_awa}(a),(b) and Figs.~\ref{fig:ad_awa}(c),(d), we can see that the misclassified examples are blurred compared with those judged correctly for reasons such as the subject being small or overlapping an object.
    Second, comparing Fig.~\ref{fig:ad_awa}(a) and Fig.~\ref{fig:ad_awa}(b), the examples in the class with the worst recall contain more unclear examples than those with the best recall.
    Examples of the giant panda class, with their black and white body color, are clearly visible in the background and are considered easy for the model to learn.
    The top classes with high recall include the giraffe, zebra, and tiger classes, easily distinguished from their subjects and backgrounds.
    Moreover, the mouse class is considered a difficult class for the model to learn because they assimilate with the background or the subject is small.
    In fact, the lower classes with lower recall include the beaver, rat, and weasel classes, whose body color is gray or whose subjects are small and easily assimilated into the background.
    Thus, classes with low baseline recall are considered to contain more opaque features that are difficult to learn.

    \begin{table}[b]
        \caption{
            Comparison of accuracy, the worst recall, the second-worst recall, and median recall when training AwA2 with CE loss, respectively, using the model shown in Table~\ref{tab:a1} and EfficientNetB2.
            Each value represents the mean and standard deviation of ten trials.
        }
        \label{tab:ad8}
        \begin{center}
            \begin{tabularx}{\linewidth}{l|CC}
                \toprule
                Metric                    & Base model shown in Table~\ref{tab:a1} & EfficientNet\cite{tan2019efficientnet} \\ \midrule
                Accuracy                  & 0.651$\pm$0.013                        & 0.717$\pm$0.005                        \\
                Recall of mouse (worst1)  & 0.118$\pm$0.051                        & 0.153$\pm$0.023                        \\
                Recall of beaver (worst2) & 0.145$\pm$0.045                        & 0.243$\pm$0.043                        \\
                Recall of moose (median)  & 0.559$\pm$0.066                        & 0.613$\pm$0.024                        \\ \bottomrule
            \end{tabularx}
        \end{center}
    \end{table}
    
    In addition, to verify if the model we used in our experiments was not expressive enough, we employed EfficientNet\cite{tan2019efficientnet} as a sufficiently expressive model.
    EfficientNet is a model with state-of-the-art reported on several datasets, including ImageNet, and its 8,515,627 parameters are larger than the 6,933,248 parameters of the model used in our experiments.
    Therefore, it can be seen as a highly expressive model.
    When EfficientNet is used, we used EfficientNetB2 with 1,000 output dimensions, which has been pretrained by ImageNet\cite{krizhevsky2012imagenet}.
    The model comprises the EfficientNetB2 output layer followed by a GAP layer, an FC layer with 512 output dimensions and activated by the ReLU function, a dropout layer with a dropout rate of 0.5, and an FC layer activated by a softmax function with 50 dimensions of the number of output dimensions of the number of classes.
    The number of epochs was set to 100 and optimized by Adam\cite{kingma2014adam}.
    The learning rate started at 0.001 and decreased by a factor of 10 every 25 epochs.
    Table~\ref{tab:ad8} shows the mean and standard deviation of accuracy and the worsts 1 and 2 and median recall over ten trials when AwA2 is used to train the base model used in Section~\ref{sec:experiment} and shown in Table~\ref{tab:a1} and EfficientNet.
    Both performance values are improved using EfficientNetB2, but the recalls of the mouse and beaver classes are still low.
    Therefore, we can say that the reason for the existence of classes with very low recall, such as the mouse and beaver classes, is not a lack of model representation.

    In summary, the reason for the low recall of the mouse and beaver classes in AwA2 is that these classes contain more examples that are unclear and difficult to learn than the classes with higher recall.
    In addition, the CNN models used in our experiments are not extremely expressive because even sufficiently expressive models cannot sufficiently improve the recall of low baseline classes that are difficult to learn.

    \section{Conclusion}
    \label{sec:conclusion}
    We proposed CAMRI loss, which improves the recall of an important class without sacrificing overall accuracy.
    First, by analyzing the contour plots of existing loss functions, we found that it is effective to reduce the angle between the feature vectors and the GT weight vectors to improve the separation of different classes in the feature space (intraclass compactness).
    To achieve this, we introduced CAMRI loss.
    Experimental results showed that CAMRI loss improved the recall of the important classes without sacrificing the accuracy compared with other methods.
    In addition, the recall was higher than that of CE loss in the setting of multiple important classes.
    We experimentally confirmed that CAMRI loss makes the intraclass compactness of the important classes relatively smaller than the other classes.

    Using CAMRI loss, we can expect to obtain a feature representation that provides a good separation of specific classes.
    Further increasing the separability of specific classes is necessary to improve recall without sacrificing accuracy.
    To achieve this, explicitly designing feature representations with the high separation of specific classes is effective, but this cannot be achieved with CAMRI loss.
    To obtain highly separable feature representations, we consider it is necessary to control for learning in layers deeper than the last FC layer of the model, which are not considered by CAMRI loss.
    Thus, developing a method to control learning in the deep layers of the model to obtain highly separable feature representations of specific classes remains our future issue.

    \section*{Acknowledgments}
    This work is partly supported by Japan science and technology agency (JST), CREST JPMJCR21D3, and Japan society for the promotion of science (JSPS), Grants-in-Aid for Scientific Research 19H04164 and 18H04099.

    \bibliographystyle{ieicetr}
    \bibliography{ref}

    \appendix

    \section{Contour plots of the loss value by related loss functions}
    \label{ap:omitplot}
    \begin{figure}[tb]
        \centerline{
            \includegraphics[width=88mm]{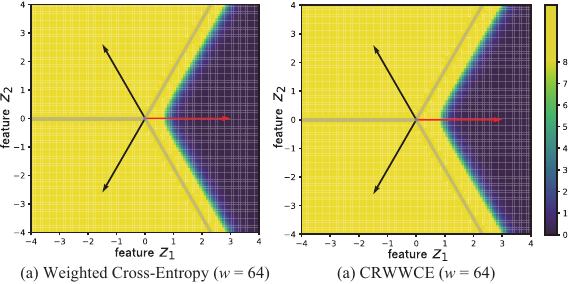}
        }
        \caption{
            The contour plots of the loss values formed by each loss function are shown.
            From left to right are WCE (the same as Fig.~\ref{fig:2}(c)) and CRWWCE loss.
            The styles of this figure are the same as in Fig.~\ref{fig:2}.
        }
        \label{fig:a}
    \end{figure}

    Fig.~\ref{fig:a} shows contour plots of loss values in the two-dimensional feature space when the loss function is WCE (left) and CRWWCE (right) loss.
    All methods form similar contour lines and do not sharpen the angle between feature vectors of the important class and GT weight vectors.
    Therefore, the separation of important classes will not be improved with these loss functions.

    \section{Model Structure Details}
    The structure of the CNN model used in our experiments with CIFAR-10 is shown in Table~\ref{tab:a1}.
    Notably, the output and input shape values change when GTSRB and AwA2 are used.
    The top of the table corresponds to the inputs, and the bottom corresponds to the outputs.
    Conv is a convolution layer, BN is a batch normalization layer, MP is a max-pooling layer, GAP is a global average pooling layer, and FC is a fully connected layer.
    The kernel size of the convolution layers is $3\times3$, the stride length is 1, and the padding is set so that the layer's output has the same length as the layer's input.
    The pooling size of max-pooling layers is $2\times2$.
    The dropout rate for dropout layers is 0.25.

    The activation function of the last FC layer, represented by $*$, depends on the loss function we used for training.
    For CE, WCE, and CRWWCE loss, it is a softmax function.
    Meanwhile, it is not activated in the case of CAMRI loss and ArcFace, because these methods calculate the loss values using the weights and teacher labels of the final FC layer.

    \label{ap:model}
    \begin{table}[t]
        \centering
        \caption{Summary of the model structure taking CIFAR-10 images as input. The activation function ($*$) of the last FC layer depends on the loss function used.
        The linear activation function passes through the output of a layer to the next layer, i.e., it does not activate.}
        \begin{tabular}[width=95mm]{@{\extracolsep{\fill}}lcc}
            \toprule
            Layer type & Output Shape          & Activation \\ \midrule
            Input      & $32\times32\times3$   & -          \\
            Conv       & $32\times32\times64$  & ReLU       \\
            Conv       & $32\times32\times64$  & ReLU       \\
            BN         & $32\times32\times64$  & -          \\
            Conv       & $32\times32\times64$  & ReLU       \\
            MP         & $16\times16\times64$  & -          \\
            Dropout    & $16\times16\times64$  & -          \\
            Conv       & $16\times16\times128$ & ReLU       \\
            Conv       & $16\times16\times128$ & ReLU       \\
            BN         & $16\times16\times128$ & -          \\
            Conv       & $16\times16\times128$ & ReLU       \\
            MP         & $8\times8\times128$   & -          \\
            Dropout    & $8\times8\times128$   & -          \\
            Conv       & $8\times8\times256$   & ReLU       \\
            Conv       & $8\times8\times256$   & ReLU       \\
            BN         & $8\times8\times256$   & -          \\
            Conv       & $8\times8\times256$   & ReLU       \\
            Conv       & $8\times8\times256$   & ReLU       \\
            Conv       & $8\times8\times256$   & ReLU       \\
            BN         & $8\times8\times256$   & -          \\
            Conv       & $8\times8\times512$   & ReLU       \\
            Conv       & $8\times8\times512$   & ReLU       \\
            GAP        & 512                   & -          \\
            FC         & 512                   & Linear     \\
            BN         & 512                   & -          \\
            FC         & 10                    & $*$        \\\bottomrule
        \end{tabular}
        \label{tab:a1}
    \end{table}

    \section{Hyperparameter values of loss functions}
    \label{ap:hypara}

    In Table \ref{tab:1}, we presented the results by tuning hyperparameters, which are $s$ and $m_k$ in CAMRI loss, $s$, and $m$ in ArcFace, $c_k^{\text{fn}}$, and $C_{k',k}^{\text{fp}}$ in CRWWCE loss, and $w_k$ in WCE loss.
    Each value of these hyperparameters is shown in Table \ref{tab:a2}.
    The top row of each cell is the recall change, meaning the difference between the recall of baseline and the recall obtained for each loss function. A positive recall change means that recall is improving.
    The row below the recall change is the value of the hyperparameter for the loss function.
    The information in each cell of Table \ref{tab:a2} corresponds to each cell in Table \ref{tab:1}.

    \section{Aggregate results considering recall only}
    \label{ap:onlyrecall}

    In Subsection~\ref{sec:eval}, Fig.~\ref{tab:1} shows the recalls in the setting where the accuracies are not decreased.
    For comparison purpose, we show the results when the recall is maximized without regarding the decrease in accuracy in Table~\ref{tab:aa1}.
    In other words, the criteria are as follows.
    \begin{enumerate}
        \item Recall and accuracy with the regular CE loss are used as the baseline.
        \item For each method, the results obtained with the parameter setting that achieves the highest recall are shown in Table~\ref{tab:aa1}.
    \end{enumerate}

    In Table~\ref{tab:aa1}, the top row of each cell is recall, the bottom row is accuracy, and the values shown are the mean and standard deviation of ten trials.
    From Table~\ref{tab:aa1}, it can be seen that it is easy to improve recall but difficult to maintain accuracy.

    \onecolumn
    \begin{table*}[t]
        \caption{
            Summary of recall changes and hyperparameter values of loss functions corresponding to the values obtained from the performance evaluation in Table \ref{tab:1}.
            Recall changes differ from the baseline results (bottom row of Table \ref{tab:1}).
            Hyperparameters are $s$ and $m_k$ in CAMRI loss, $s$, and $m$ in ArcFace, $c_k^{\text{fn}}$, and $C_{k',k}^{\text{fp}}$ in CRWWCE loss, and $w_k$ in WCE loss.
            Recall change and hyperparameter values are shown in each cell from the top row.}
        \begin{center}
            \begin{tabularx}{\textwidth}{@{}lc*{9}{C}@{}}
                \toprule
                & & \multicolumn{3}{c}{CIFAR-10} & ~ & GTSRB & ~ & ~ & AwA2 & ~ \\
                \cmidrule(lr){3-5} \cmidrule(lr){6-8} \cmidrule(lr){9-11}
                ~       & ~                                       & cat       & dog       & airplane  & limit-80-end & no-passing-end & limit-80 & mouse     & beaver    & moose     \\
                method  & ~                                       & (worst1)  & (worst2)  & (median)  & (worst1)     & (worst2)       & (median) & (worst1)  & (worst2) & (median) \\\midrule
                ~       & Recall change                           & 0.054     & 0.016     & 0.029     & 0.075        & 0.09           & 0.006    & 0.011     & 0.093     & 0.086     \\
                CAMRI   & $s$                                     & 16        & 16        & 8         & 64           & 4              & 64       & 16        & 16        & 16        \\
                ~       & $m_k$                                   & $4\pi/64$ & $5\pi/64$ & $4\pi/64$ & $\pi/64$     & $8\pi/64$      & $\pi/64$ & $8\pi/64$ & $6\pi/64$ & $5\pi/64$ \\ \midrule
                ~       & Recall change                           & 0.027     & 0.008     & 0.022     & 0.048        & 0.071          & 0.004    & -0.017    & 0.042     & 0.053     \\
                ArcFace & $s$                                     & 4         & 4         & 8         & 32           & 32             & 32       & 16        & 8         & 4         \\
                ~       & $m$                                     & $4\pi/64$ & $\pi/64$  & 0         & $3\pi/64$    & $4\pi/64$      & $\pi/64$ & $\pi/64$  & $5\pi/64$ & $5\pi/64$ \\ \midrule

                CRWWCE  & Recall change                           & -         & -         & -         & 0.036        & 0.053          & -        & -         & -         & -         \\
                ~       & $c_k^{\text{fn}}, C_{k',k}^{\text{fp}}$ & -         & -         & -         & 4            & 8              & -        & -         & -         & -         \\ \midrule
                WCE     & Recall change                           & 0.064     & -         & -         & 0.069        & 0.075          & -        & -         & -         & -         \\
                ~       & $w_k$                                   & 4         & -         & -         & 24           & 24             & -        & -         & -         & -         \\  \bottomrule
            \end{tabularx}
            \label{tab:a2}
        \end{center}
    \end{table*}
    \begin{table*}[tb]
        \caption{
            Mean and standard deviation of recall of the important class, the upper row, and accuracy, the lower row, over ten trials.
            The results presented here show the cases with the highest recall for each method.}
        \begin{center}
            \footnotesize
            \scalebox{0.86}{
                \begin{tabularx}{1.15\textwidth}{@{}ll*{8}{C}c@{}}
                    \toprule
                    &          &                 & CIFAR-10        &                 &                 & GTSRB           &                 &                 & AwA2            &                 \\
                    \cmidrule(lr){3-5} \cmidrule(lr){6-8} \cmidrule(lr){9-11}
                    &          & cat             & dog             & airplane        & limit-80-end    & no-passing-end  & limit-80        & mouse           & beaver          & moose           \\
                    method     & metric          & (worst1)        & (worst2)        & (median)        & (worst1)        & (worst2)        & (median)        & (worst1)        & (worst2)        & (median)        \\ \midrule
                    CAMRI      & Recall   & 0.865$\pm$0.023 & 0.920$\pm$0.014 & 0.920$\pm$0.018 & 0.843$\pm$0.055 & 0.967$\pm$0.028 & 0.995$\pm$0.003 & 0.130$\pm$0.046 & 0.238$\pm$0.053 & 0.646$\pm$0.054 \\

                    (proposal) & Accuracy & 0.720$\pm$0.006 & 0.727$\pm$0.004 & 0.650$\pm$0.036 & 0.874$\pm$0.030 & 0.890$\pm$0.022 & 0.981$\pm$0.001 & 0.667$\pm$0.014 & 0.664$\pm$0.015 & 0.656$\pm$0.025 \\   \midrule
                    ArcFace    & Recall   & 0.766$\pm$0.041 & 0.843$\pm$0.032 & 0.911$\pm$0.029 & 0.843$\pm$0.043 & 0.942$\pm$0.070 & 0.994$\pm$0.003 & 0.300$\pm$0.458 & 0.188$\pm$0.052 & 0.612$\pm$0.046 \\

                    ~          & Accuracy & 0.880$\pm$0.004 & 0.880$\pm$0.006 & 0.880$\pm$0.005 & 0.528$\pm$0.063 & 0.982$\pm$0.002 & 0.982$\pm$0.003 & 0.008$\pm$0.006 & 0.666$\pm$0.009 & 0.677$\pm$0.009 \\ \midrule

                    CRWWCE     & Recall   & 0.764$\pm$0.018 & 0.829$\pm$0.036 & 0.901$\pm$0.023 & 0.823$\pm$0.028 & 0.933$\pm$0.044 & 0.992$\pm$0.003 & 0.094$\pm$0.047 & 0.174$\pm$0.082 & 0.548$\pm$0.065 \\
                    ~          & Accuracy & 0.876$\pm$0.004 & 0.877$\pm$0.005 & 0.869$\pm$0.007 & 0.978$\pm$0.004 & 0.979$\pm$0.002 & 0.978$\pm$0.003 & 0.640$\pm$0.014 & 0.640$\pm$0.014 & 0.640$\pm$0.014 \\\midrule
                    WCE        & Recall   & 0.825$\pm$0.021 & 0.897$\pm$0.025 & 0.937$\pm$0.014 & 0.831$\pm$0.024 & 0.945$\pm$0.039 & 0.995$\pm$0.004 & 0.137$\pm$0.075 & 0.203$\pm$0.082 & 0.571$\pm$0.071 \\
                    ~          & Accuracy & 0.863$\pm$0.006 & 0.848$\pm$0.013 & 0.872$\pm$0.004 & 0.979$\pm$0.004 & 0.981$\pm$0.005 & 0.980$\pm$0.004 & 0.522$\pm$0.056 & 0.614$\pm$0.012 & 0.616$\pm$0.011 \\ \midrule
                    CE & Recall & 0.738$\pm$0.041 & 0.834$\pm$0.030 & 0.888$\pm$0.022 & 0.760$\pm$0.049 & 0.870$\pm$0.097 & 0.989$\pm$0.003  & 0.118$\pm$0.051 &
                    0.145$\pm$0.045 & 0.559$\pm$0.066 \\
                    (baseline) & Accuracy & 0.879$\pm$0.005 & 0.879$\pm$0.005 & 0.879$\pm$0.005 & 0.980$\pm$0.002 & 0.980$\pm$0.002 & 0.980$\pm$0.002  & 0.651$\pm$0.013 & 0.651$\pm$0.013 & 0.651$\pm$0.013 \\ \bottomrule
                \end{tabularx}}
            \label{tab:aa1}
        \end{center}
    \end{table*}
    \twocolumn

    \profile[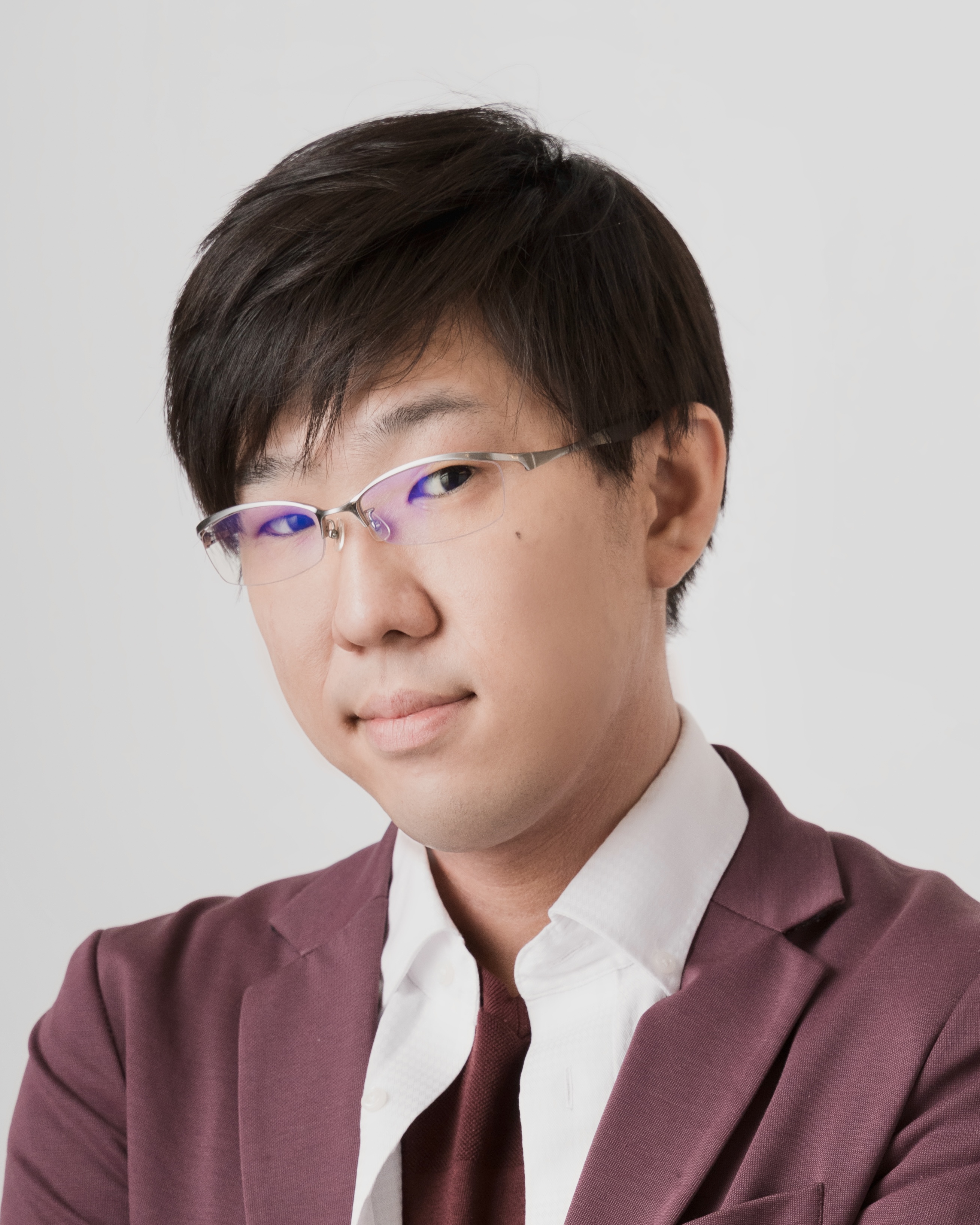]{Daiki Nishiyama}{received the B.S. degree in information engineering from University of Tsukuba, Japan, in 2021. Since 2021, he has been a Graduate Student with Master's Program in Computer Science, University of Tsukuba, Japan, and also a research part-time worker with the Center for Advanced Intelligence Projects, RIKEN. His research interests include machine learning, data mining, explainable AI, and their application.}
    \profile[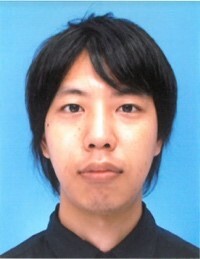]{Kazuto Fukuchi}{received a Ph.D. degree from University of Tsukuba, Japan, in 2018. Since 2019, he has been an Assistant Professor with the Faculty of Engineering, Information and Systems, University of Tsukuba, Japan, and also a Visiting Researcher with the Center for Advanced Intelligence Project, RIKEN, Japan. His research interests include mathematical statistics, machine learning, and their application.}
    \profile[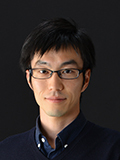]{Youhei Akimoto}{received the B.S. degree in computer science in 2007, and the M.S. and Ph.D. degrees in computational intelligence and systems science from Tokyo Institute of Technology, Japan, in 2008 and 2011, respectively. From 2010 to 2011, he was a Research Fellow of JSPS in Japan, and from 2011 to 2013, he was a Post-Doctoral Research Fellow with INRIA in France. From 2013 to 2018, he was an Assistant Professor with Shinshu University, Japan. Since 2018, he has been an Associate Professor with University of Tsukuba, Japan as well as a Visiting Researcher with the Center for Advanced Intelligence Projects, RIKEN. He served as a Track Chair for the continuous optimization track of GECCO in 2015 and 2016. He is an Associate Editor of ACM TELO and is on the editorial board of the ECJ. He won the Best Paper Award at GECCO 2018 and FOGA 2019. His research interests include design principles, theoretical analyses, and applications of stochastic search heuristics.}
    \newpage
    \profile[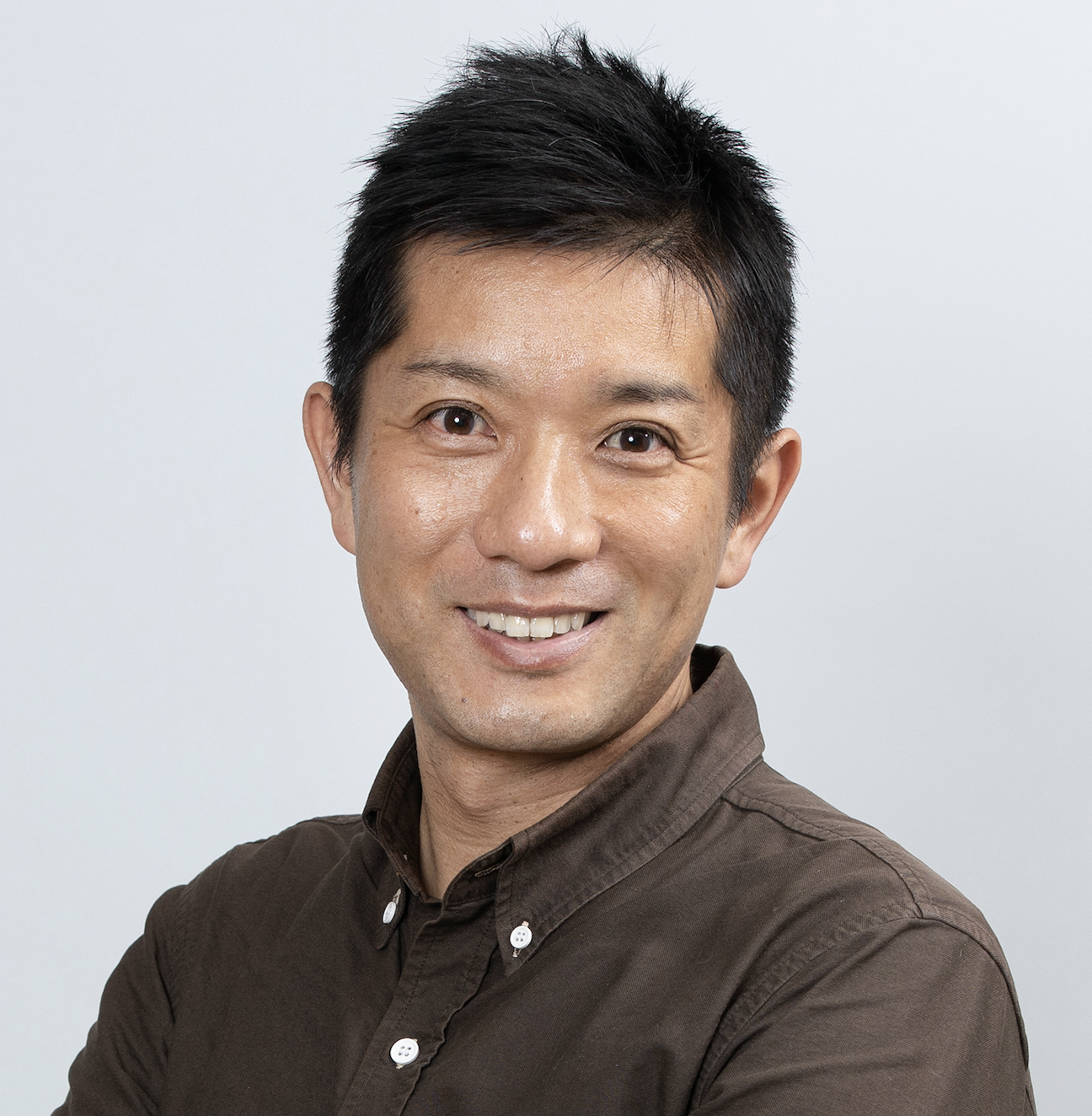]{Jun Sakuma}{received a Ph.D. degree in Engineering from the Tokyo Institute of Technology, Tokyo Japan in 2003. He has been a professor in the Department of Computer Science, School of System and Information Engineering, University of Tsukuba, Tsukuba, Japan, since 2016. He has also been a team leader of Artificial Intelligence Security and Privacy team in the Center for Advanced Intelligence Project, RIKEN, since 2016. Prior to that, he worked as an associate professor in the same department of University of Tsukuba (2009--2016). He worked as an assistant professor in the Department of Computational Intelligence and Systems Science, Interdisciplinary Graduate School of Science and Engineering, Tokyo Institute of Technology, Tokyo, Japan (2004--2009). He worked as a researcher at Tokyo Research Laboratory, IBM, Tokyo Japan (2003--2004). His research interests are machine learning, data privacy and AI security. He is a member of the Institute of Electronics, Information and Communication Engineers of Japan (IEICE).}
\end{document}